\documentclass[a4paper,fleqn]{cas-dc}

\usepackage[numbers,sort&compress]{natbib}
\usepackage{placeins}

\usepackage{xcolor}
\newcommand{\rev}[1]{\textcolor{black}{#1}}

\begin{document}
\let\WriteBookmarks\relax
\def\floatpagepagefraction{1}
\def\textpagefraction{.001}

\shorttitle{Compositional Benchmark Synthesis for Hierarchical Action Recognition}

\shortauthors{Soleimani et al.}

\title[mode = title]{Compositional Benchmark Synthesis for Hierarchical Human Action Recognition}

\author[1]{Farnaz Soleimani}[orcid=0009-0008-8960-4869]
\cormark[1]
\ead{farnaz.soleimani@u-pec.fr}
\credit{Conceptualization, Methodology, Software, Data curation, Investigation, Formal analysis, Visualization, Validation, Writing, original draft}

\author[1]{Abdelghani Chibani}
\ead{abdelghani.chibani@u-pec.fr}
\credit{Writing, review and editing}

\author[1]{Yacine Amirat}
\ead{amirat@u-pec.fr}
\credit{Writing, review and editing}

\author[1]{Ghazaleh Khodabandelou}
\ead{ghazaleh.khodabandelou@u-pec.fr}
\credit{Conceptualization, Methodology, Supervision, Writing, review and editing}

\affiliation[1]{
    organization={LISSI Laboratory, University of Paris-Est Créteil (UPEC)},
    addressline={122 Rue Paul Armangot},
    city={Vitry-sur-Seine},
    postcode={94400},
    country={France}
}

\cortext[1]{Corresponding author}

\begin{abstract}
Recognizing human behavior across levels of abstraction, from atomic
actions to long-horizon intentions, requires data annotated along a
semantic hierarchy. Large corpora provide isolated, atomically labeled
clips without temporal composition, whereas recorded composite-activity
corpora offer shallow, domain-narrow, fixed hierarchies. A
benchmark-generation and evaluation framework is proposed that
synthesizes a four-level hierarchical-intention benchmark, spanning
actions, activities, low-level intentions (LLIs), and high-level
intentions (HLIs), from a flat single-label action corpus while
retaining real pre-extracted features at the action level. Episodes are
assembled by a transition model under a subject-consistency constraint,
and a coverage-aware sampler reduces the subject-usage Gini from $0.566$
to $0.248$. \rev{Synthesizing such a benchmark raises a circular-supervision risk that recorded datasets avoid: if the rules generating the episodes also govern the evaluation, models can succeed by recovering the generator rather than through genuine reasoning.} Validity is
addressed by design, holding sequence-generation rules disjoint from the
first-order-logic rules used at evaluation. The instantiation yields $15{,}002$ episodes.
\rev{Four reference baselines from different model families characterize
difficulty, not as recognition methods. A compositional held-out gap of
$0.13$ to $0.17$ macro-F1 appears across all baselines, including a
graph-aware model that recognizes best yet does not close the gap,
indicating a structural property of the benchmark rather than a
model artifact. A logic-free baseline still violates the held-out
semantic rules above their intrinsic data rate, and the order-destroying
control changes macro-F1 within seed variation, serving as a
generator-consistency check.} The ontology, transition model, and
generator are released so the benchmark can be regenerated and extended.
\end{abstract}

\begin{graphicalabstract}
\includegraphics[width=\textwidth]{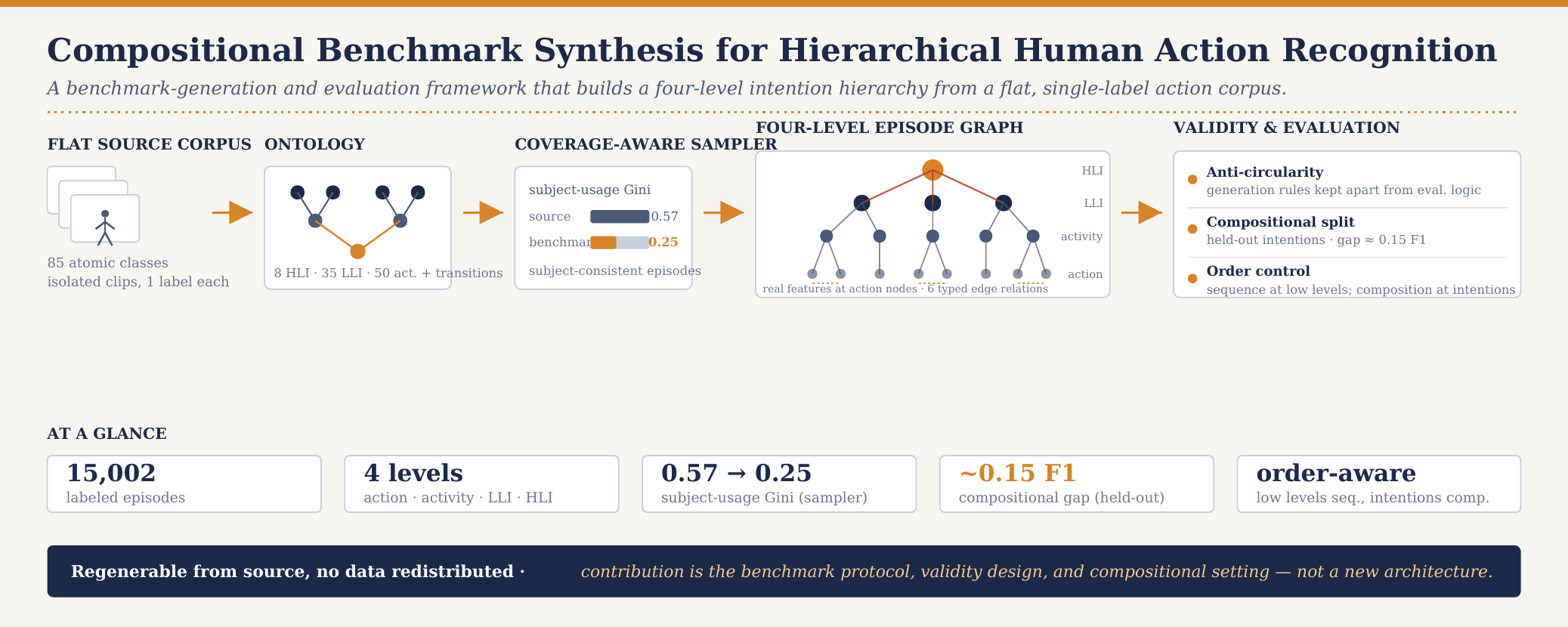}
\end{graphicalabstract}

\begin{highlights}
\item Framework synthesizes a 4-level intention benchmark from flat action corpora
\item Coverage-aware sampler cuts subject-usage Gini from 0.57 to 0.25
\item Generation rules kept disjoint from evaluation logic to avoid circular supervision
\item \rev{Order-destroying control changes macro-F1 within seed noise, a generator-consistency check}
\item \rev{Compositional gap of 0.13 to 0.17 macro-F1 holds across four baseline families including a graph model, showing it is structural}
\end{highlights}


\maketitle

\section{Introduction}\label{sec:intro}

Recognizing human behavior at multiple abstraction levels underpins applications from assistive robotics to human--robot collaboration. In these settings, inferring the instant atomic action (\emph{what}) is insufficient; the long-horizon intention (\emph{why}) driving the sequence must also be inferred. This reasoning organizes naturally into a semantic hierarchy: atomic actions compose into activities, activities express short-horizon intentions, and these form behavioral episodes. While flat action classification has matured, it fails to exercise the compositional and hierarchical reasoning required at higher levels.

Progress on multi-level reasoning remains constrained by data fragmentation. Large single-label corpora \cite{liu2019ntu} provide scale but lack structure above isolated clips. Conversely, recorded composite corpora provide temporal segmentation \cite{sener2022assembly101}, hierarchical daily-living annotations \cite{benavent2025enhancing}, or multi-tier localization \cite{liu2022fineaction}, but their hierarchies remain shallow, domain-narrow, and fixed. No existing resource pairs a deep intention hierarchy with the scale and subject structure of a large single-label corpus.

Instead of introducing a newly recorded dataset, this work proposes a framework to synthesize hierarchical benchmarks from existing flat action corpora by composing atomic clips, an approach grounded in grammar-based models \cite{qi2017predicting} and synthetic data generation \cite{8099761}. However, synthesis introduces a critical \textit{circular supervision} risk absent from recorded data: if generation rules also define evaluation, models can shortcut the task by recovering the generator rather than reasoning \cite{gill2025has}. Cross-subject protocols further complicate synthesis; enforcing performer consistency across disjoint subject subsets makes sampling non-uniform, causing naive generation to collapse either diversity or structural representation.

In response, we propose a framework that constructs a four-level hierarchical-intention benchmark while addressing validity and coverage by design. The protocol retains real pre-extracted features at the action level and assembles episodes using a transition model with a coverage-aware sampler under strict subject consistency. To prevent circularity, sequence-generation rules remain disjoint from evaluation-time first-order-logic constraints. Finally, compositional held-out partitions at the intention level are combined with subject-disjoint splits to separate generalization from memorization \cite{materzynska2020something}. 

\rev{To characterize benchmark difficulty, we evaluate four baselines from distinct model families (a hierarchical transformer, a sequential transformer, an order-free bag-of-actions model, and a relational graph network over the typed episode graph) using ordinary cross-entropy without logical constraints. The evaluation logs recognition accuracy, an order-destroying control, and first-order-logic violation rates. Crucially, this paper does not propose a new recognition architecture; the central contribution is a novel benchmark-construction and validation protocol for hierarchical pattern recognition.}

\textbf{Novelty and contributions.}
The novelty lies in the \emph{joint protocol}: converting a flat,
isolated-clip action corpus into a four-level hierarchical intention
benchmark while simultaneously enforcing subject consistency, coverage
balancing, disjoint generation and evaluation rules, and compositional
held-out testing. To the best of our knowledge these elements have not
previously been integrated into a single benchmark framework for
hierarchical human action recognition. The contributions are
(i) a regenerable synthesis protocol that preserves real action-level
features while adding activity, LLI, and HLI structure;
(ii) a coverage-aware subject sampler that reduces performer imbalance
under cross-subject constraints;
(iii) an anti-circularity validation design separating the generator
from evaluation-time semantic rules;
(iv) a compositional held-out split at the LLI--HLI association level;
and (v) an empirical protocol combining recognition, logical-violation
rates, and order-destroying controls. The framework is instantiated on a
skeleton action corpus, producing $15{,}002$ episodes; the ontology,
transition model, and generator are released so that the benchmark can
be reconstructed and extended under the source corpus's native license
rather than redistributed directly.
 
\section{Related Work}
\label{sec:related}
\paragraph{Composite and Procedural Activity Datasets}
Long, structured human behavior is primarily studied on \emph{recorded} corpora (e.g., Breakfast, 50~Salads, GTEA, Assembly101 \cite{sener2022assembly101}, FineAction \cite{liu2022fineaction}). Despite recent additions of hierarchical daily-living annotations \cite{benavent2025enhancing}, these resources share three limitations: their hierarchies are shallow (typically two levels instead of four reaching intention), they are confined to narrow domains, and their recorded sequences are fixed. \rev{Table~\ref{tbl:datasets} summarizes these structural differences along our core evaluation axes.}

\begin{table}[!htbp]
\color{black}
\caption{\rev{Composite- and procedural-activity datasets compared along
the axes relevant to this work. ``Levels'' counts annotated semantic
tiers; ``Label types'' lists the highest semantic label available;
``Regen.'' indicates whether the resource can be regenerated or
recombined; ``Comp.\ split'' indicates a held-out compositional
partition. No existing resource reaches the intention level with
a regenerable, compositionally-splittable structure.}}\label{tbl:datasets}
\small
\setlength{\tabcolsep}{3pt}
\begin{tabular*}{\tblwidth}{@{}LCLCCC@{}}
\toprule
Dataset & Levels & Label types & Regen. & \parbox{1.2cm}{\centering Comp.\ split} \\
\midrule
Breakfast        & 2 & Action, activity        & No  & No  \\
50~Salads        & 2 & Action, activity        & No  & No  \\
GTEA             & 2 & Action, activity        & No  & No  \\
Assembly101      & 2 & Action, coarse segment  & No  & No  \\
Hierarchical ADL & 2 & Activity (coarse/fine)  & No  & No  \\
FineAction       & 3 & Action (3 granularities)& No  & No  \\
\textbf{This work} & \textbf{4} & \textbf{Action, activity, LLI, HLI} & \textbf{Yes} & \textbf{Yes} \\
\bottomrule
\end{tabular*}
\end{table}
 
\paragraph{Synthetic and Generated Activity Data}
Synthetic generation often alleviates recorded data scarcity by rendering photo-realistic clips \cite{8099761} or virtual household activities \cite{ozdel2024gaze}. While these synthesize visual \emph{appearance} or \emph{motion}, our work differs fundamentally: we recombine real pre-extracted skeleton features into a multi-level semantic \emph{label structure}, targeting compositional hierarchy rather than visual realism.
 
\paragraph{Grammar-Based and Compositional Activity Modeling}
Compositional behavior is widely captured via generative grammars, And--Or graphs \cite{qi2017predicting}, adversarial rules \cite{piergiovanni2020adversarial}, and probabilistic automata \cite{cruz2020inferring, todorovic2012human}. While we adopt a transition model from this family, it serves purely as a controllable \emph{benchmark generator} whose rules are kept strictly disjoint from evaluation constraints, rather than acting as a recognizer.
 
\paragraph{Intention and Goal-Conditioned Anticipation}
Modeling intention typically focuses on anticipation \cite{mascaro2023intention, zatsarynna2023action, cao2025vision}. These works establish intention as a useful supervisory signal but operate on shallow, two-level egocentric hierarchies. They do not address the four-level, graph-structured hierarchy and typed logical validation proposed here.
 
\paragraph{Neuro-Symbolic and Logic-Constrained Recognition}
Injecting symbolic knowledge into neural training is well-established, utilizing differentiable penalties \cite{xu2018semantic}, logic tensors \cite{badreddine2022logic}, mean-field inference \cite{xu2024logicmp}, and hierarchical constraints \cite{ye2025zero}. We utilize standard logical machinery, but uniquely deploy typed constraints strictly for \emph{evaluation-time validation}—measuring benchmark anti-circularity via violation rates—rather than relying on them to artificially boost model accuracy.
 
\paragraph{Heterogeneous Graph Transformers}
Graph- and transformer-based encoders are standard for structured network and skeleton-based recognition \cite{hu2020heterogeneous, zhu2025hhgt, yan2026skelformer}. \rev{Our benchmark is encoder-agnostic (Section~\ref{sec:framework}). To characterize structural difficulty without proposing a new architecture, our evaluation relies on generic baselines across model families (a hierarchical transformer, sequential transformer, bag-of-actions MLP, and relational graph network), ensuring the reported difficulty reflects dataset properties rather than backbone specifics.}
 
\paragraph{Compositional Generalization and Benchmark Validity}
Holding out unseen combinations is standard for probing compositional generalization \cite{materzynska2020something, li2024c2c, kim2020cogs}. We apply this at the intention level, alongside subject-disjoint splits, to separate generalization from memorization. Furthermore, synthetic evaluations risk shortcut learning \cite{gill2025has}. We operationalize benchmark validity \cite{gebru2021datasheets} for hierarchical activity via a novel anti-circularity protocol: separating generation from evaluation rules, introducing order controls, and tracking semantic violations.
\section{Hierarchical Ontology}\label{sec:ontology}

Behavior is organized over four typed levels. An \emph{action} is an
atomic, single-label clip from the source corpus; an \emph{activity} is
a short composition of actions; a \emph{low-level intention} (LLI) is a  
goal-directed unit of activities; and a \emph{high-level intention}
(HLI) is a long-horizon episode of LLIs. The instantiated ontology
comprises $8$ HLIs, $35$ LLIs, $50$ activities, and $85$ source action
classes (Table~\ref{tbl:ontology}); the catalog is released with the
generator.

\begin{table}[!htbp]
\caption{The instantiated four-level ontology.}\label{tbl:ontology}
\begin{tabular*}{\tblwidth}{@{}LLL@{}}
\toprule
Level & \# Categories & Composed of \\
\midrule
High-level intention (HLI) & 8  & 2--5 LLIs \\
Low-level intention (LLI)  & 35 & 2--4 activities \\
Activity                   & 50 & 1--2 actions \\
Action                     & 85 & atomic source clips \\
\bottomrule
\end{tabular*}
\end{table}

The ontology follows four principles. Semantic realism governs
grouping: actions are grouped only when the grouping reflects a
coherent behavior. Full coverage is required: every source action class
appears in at least one activity. Activities are modality-agnostic, so
further modalities attach without revising the ontology. Finally,
templates are flexible: each parent declares a candidate pool and the
generator samples a subset per occurrence, which keeps
subject-eligibility pools healthy.

\textbf{Multi-parent LLIs.}
Four situational LLIs (a brief drink break, a phone check, a walking
transit, and a seated rest) are attached to several HLIs. This
multi-parent structure renders the compositional held-out evaluation
(Section~\ref{sec:validity}) non-degenerate: a single-parent ontology
would reduce the compositional test to memorization. The remaining LLIs
stay single-parent where semantically appropriate.
\begin{figure*}[t]
  \centering
  \includegraphics[width=\textwidth]{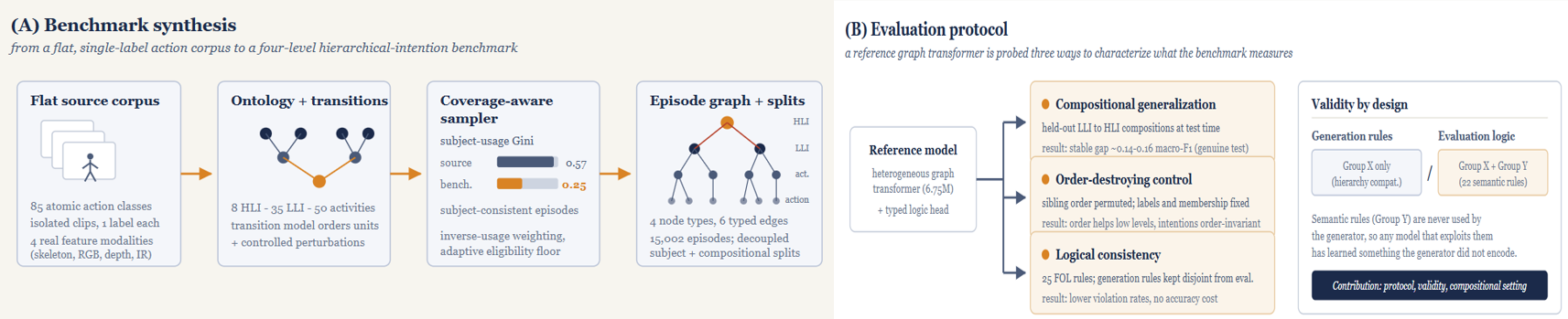}
  \caption{Overview of the experiment. (A) Benchmark synthesis: a flat,
  single-label action corpus is composed into four-level
  hierarchical-intention episodes through an ontology and transition
  model, a coverage-aware subject-consistent sampler that reduces the
  subject-usage Gini from $0.566$ to $0.248$, and a heterogeneous-graph
  representation with decoupled subject-disjoint and compositional
  splits. (B) Evaluation protocol: \rev{reference baselines from
  different model families are} probed for compositional generalization,
  sensitivity to temporal order, and \rev{first-order-logic violation
  rates}. Validity is enforced by holding the
  generation rules (Group~X only) disjoint from the evaluation logic
  (Group~X plus the 22 semantic Group~Y rules), so that exploiting the
  semantic rules reflects learning the generator did not encode.}
  \label{fig:pipeline}
\end{figure*}
\section{Synthesis Framework}\label{sec:framework}

Each episode is a tree of typed nodes rooted at one HLI, expanded
recursively: an HLI samples $2$--$5$ LLIs, each LLI samples $2$--$4$
activities, and each activity samples $1$--$2$ source action clips. The
novel aspect of the synthesis is the combination of real action-level
features with generated higher-level semantic structure, so that models
are evaluated on hierarchical pattern recognition without requiring a new
recording campaign. The
pipeline links source features, the ontology and transition model, the
coverage-aware subject sampler, heterogeneous-graph episode construction,
and decoupled split generation.

\rev{\textbf{Action-level features.}
Each source action clip is represented by features pre-extracted once
and thereafter referenced rather than recomputed. Skeleton sequences
($3$ channels, up to $300$ frames, $25$ joints, $2$ bodies) are encoded
with a PoseC3D SlowOnly-R50 backbone pretrained on the NTU120
cross-subject split into a $2048$-dimensional descriptor, and the RGB,
depth, and infrared streams each with a VideoMAE-family extractor into a
$1024$-dimensional descriptor. These four vectors define every action
node; higher-level nodes carry no raw features and aggregate their
children. Because descriptors are referenced by clip key, episodes
remain lightweight and no source data are duplicated.}

\textbf{Transition-governed generation.}
Rather than concatenating clips at random, the ordering of units within
each parent follows a transition model that encodes preferred temporal
orderings declared in the ontology (e.g.\ approach precedes greeting)
as soft constraints, and controlled label-preserving perturbations
(activity substitution, order perturbation, and filler insertion, each
fired per episode with probability $0.1$) inject realistic variation.

\textbf{Subject-consistent, coverage-aware sampling.}
All clips in an episode share one performer, preserving motion
continuity. For each HLI template, the performers able to instantiate at
least two LLIs (each with at least two viable activities) form the
eligible pool; pools range from $40$ to $60$ of the $60$ source
performers. Because source coverage is non-uniform, performers are
sampled inversely to prior usage, and the constraint relaxes adaptively
to a floor of $2$ eligible performers when needed.

\textbf{Heterogeneous graph construction.}
Each episode is materialized as a heterogeneous graph with four node
types and six typed edge relations: three compositional
(action $\rightarrow$ activity, activity $\rightarrow$ LLI,
LLI $\rightarrow$ HLI) and three temporal (\textsc{next} among siblings at
the action, activity, and LLI levels); reverse edges are added inside
the model for top-down message passing. Action nodes carry the real
pre-extracted features, referenced by source clip key (which also
encodes the performer), so features are not copied into the graphs.

\textbf{Decoupled splitting.}
Synthesis writes episodes with unique identifiers and no split
assignment; a separate procedure produces the splits, so split ratios
and held-out compositions can be revised without regenerating data.

\section{Validity by Design}\label{sec:validity}

A synthesized benchmark whose generator, labels, and evaluation rules
coincide can be solved by recovering the generator. The novel validity
principle in this work is to make this shortcut testable: generation,
logic-based evaluation, compositional splitting, and order destruction are
kept as separate components. Two mechanisms guard against circular
supervision.

\textbf{Disjoint generation and evaluation rules.}
Evaluation uses $25$ typed first-order-logic rules across the four
levels (Table~\ref{tbl:fol}), in two groups. Group~X ($3$ rules) are
hierarchy-compatibility rules the generator also enforces, limited to
cross-level consistency. Group~Y ($22$ rules) are semantic constraints
the generator does \emph{not} use (temporal precedence, mutual
exclusion, cardinality, and co-occurrence implications) that any
annotator would apply independently of how the data were produced.
\rev{Because the generator never enforces Group~Y, the rate at which an
ordinary logic-free model violates these rules is a property of the
benchmark rather than of any model: a non-trivial Group~Y rate shows the
data are not already rule-consistent by construction, so the benchmark
cannot be solved by recovering the generator (Section~\ref{sec:results}).
The rule set is included because one design goal is to support
neuro-symbolic reasoning; here it serves only as an evaluation-time
validity instrument, and integrating logical constraints into training
is left to future work.}

\begin{table}[!htbp]
\caption{First-order-logic rules by level and group. Group~X is shared
with the generator; Group~Y is semantic and evaluation-only.}\label{tbl:fol}
\begin{tabular*}{\tblwidth}{@{}LLLL@{}}
\toprule
Level & \# Rules & Group~X & Group~Y \\
\midrule
Action   & 6  & 1 & 5 \\
Activity & 7  & 1 & 6 \\
LLI      & 8  & 1 & 7 \\
HLI      & 4  & 0 & 4 \\
\midrule
Total    & 25 & 3 & 22 \\
\bottomrule
\end{tabular*}
\end{table}

\textbf{Compositional held-out protocol.}
Beyond subject-disjoint partitioning, one parent association is
withheld from training for each multi-parent LLI, so the concept must
transfer to an unseen high-level context at test time. This yields a
dedicated compositional test split alongside the standard split.

\textbf{Order-destroying control.}
To test how much of the benchmark's signal depends on hierarchical
\emph{temporal order}, a matched control is constructed by taking each
synthesized episode and permuting only the order of its siblings (LLIs
within the episode, activities within each LLI, actions within each
activity) while holding every label, performer, clip, length, and
parent--child membership fixed. The six typed edges are rebuilt over the
shuffled order. A model that still scores well on the control is
recovering label co-occurrence rather than sequence. Because short
episodes contain unshuffleable single-child parents, the control is
conservative: order was altered in $96\%$ of episodes at the action
level and $71\%$ at the LLI level.

\textbf{Scope of validation.}
The present validation focuses on structural coverage, compositional generalization, order sensitivity, and logical consistency. \rev{As a lightweight plausibility check, five independent annotators judged whether the action sequences of $150$ sampled episodes form coherent
behavior; the majority-vote plausibility rate was $0.86$ ($43/50$) with Fleiss' $\kappa = 0.725$ (substantial agreement), indicating that the synthesized episodes are generally judged plausible.}

\section{Benchmark Statistics}\label{sec:stats}

The instantiated benchmark contains $15{,}002$ episodes with $330{,}145$
forward typed edges (full counts in Table~\ref{tbl:counts}). Episodes
average $6.95$ actions (median $6$, range $4$--$17$), $2.80$ LLIs, and
$6.30$ activities; each LLI averages $2.25$ activities and each activity
$1.10$ actions. The eight HLIs are populated nearly uniformly ($2{,}188$
episodes for the two largest versus $1{,}771$ for the rest).

\begin{table}[!htbp]
\caption{Benchmark composition (instantiated from a skeleton corpus).}\label{tbl:counts}
\small
\setlength{\tabcolsep}{3pt}
\begin{tabular}{@{}p{0.60\columnwidth}p{0.32\columnwidth}@{}}
\toprule
Quantity & Value \\
\midrule
Episodes (HLI samples) & 15{,}002 \\
\midrule
Node counts & \\
\quad Action nodes & 104{,}310 \\
\quad Activity nodes & 94{,}455 \\
\quad LLI nodes & 42{,}072 \\
\quad HLI nodes & 15{,}002 \\
\midrule
Edge counts & \\
\quad Compositional edges & 240{,}837 \\
\quad Temporal edges & 89{,}308 \\
\quad Total forward typed edges & 330{,}145 \\
\midrule
Distinct performers used & 60 \\
Mean / median episode length & 6.95 / 6 actions \\
Source availability Gini & 0.566 \\
Realized subject-usage Gini & 0.248 \\
\bottomrule
\end{tabular}
\end{table}

\textbf{Coverage and the sampler.}
The source corpus is markedly skewed: across $60$ performers the mean
coverage is $44.7$ of $85$ classes (range $32$--$85$), a clip-availability
Gini of $0.566$. The coverage-aware sampler reduces the realized
subject-usage Gini to $0.248$ (most-used performer in $364$ episodes,
least in $64$). Eligible-subject pools per HLI range from $40$ to $60$,
reflecting that medical, social, and exercise templates demand rarer
source classes.

\begin{figure*}[!t]
  \centering
  \begin{minipage}[t]{0.31\textwidth}
    \centering
    \includegraphics[width=0.9\linewidth]{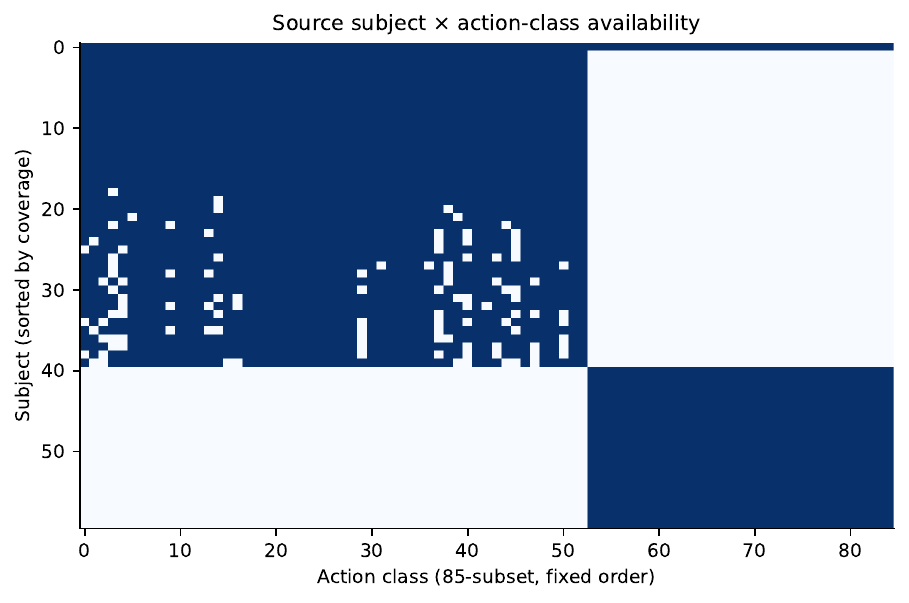}\par
    \vspace{1mm}\footnotesize (a) Subject--action availability.
  \end{minipage}\hfill
  \begin{minipage}[t]{0.31\textwidth}
    \centering
    \includegraphics[width=0.8\linewidth]{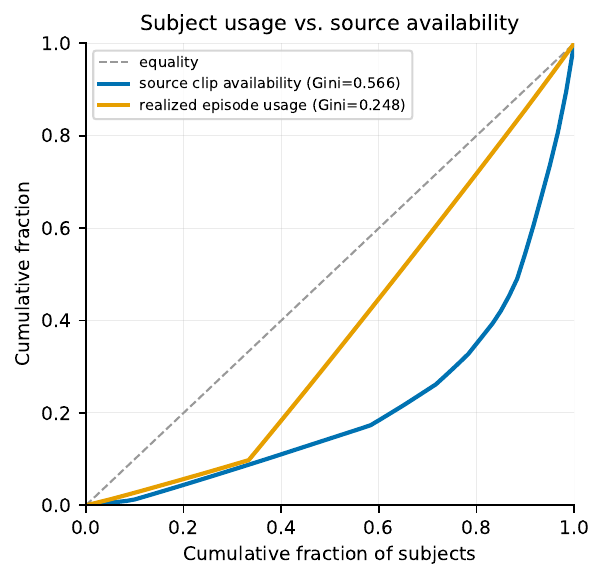}\par
    \vspace{1mm}\footnotesize (b) Source and realized usage Gini.
  \end{minipage}\hfill
  \begin{minipage}[t]{0.31\textwidth}
    \centering
    \includegraphics[width=1.1\linewidth]{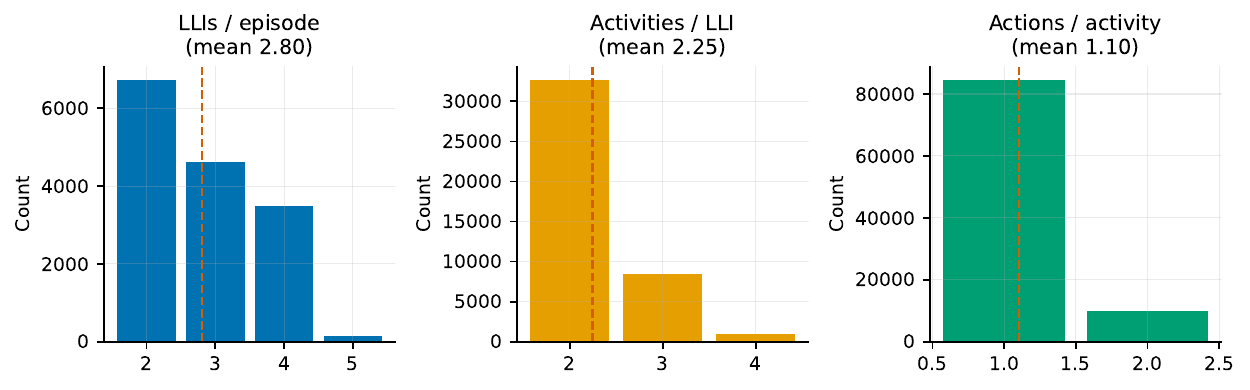}\par
    \vspace{1mm}\footnotesize (c) Branching distributions.
  \end{minipage}
  \caption{Compact benchmark statistics. The panels show (a) source subject-by-class availability,
  (b) Lorenz curves of source availability versus realized subject usage, and
  (c) per-level branching distributions of the synthesized episodes.}
  \label{fig:stats}
\end{figure*}

\textbf{Structure and long tail.}
Structural statistics are summarized in Fig.~\ref{fig:stats}. The action
level is long-tailed: the most frequent class supplies $10{,}996$ nodes
whereas the rarest dozen supply fewer than $30$ each. \rev{This tail
bounds macro-averaged recognition: across all four baselines the action
level yields the lowest per-level macro-F1 (Table~\ref{tbl:recog}),
consistent with the rare-class tail limiting action-level performance
independently of architecture.}

\section{Reference Evaluation}\label{sec:results}

\rev{Four reference baselines from different model families
characterize benchmark difficulty; none is a methodological
contribution, and all use ordinary cross-entropy without logical
constraints. The primary baseline is a non-graph hierarchical
transformer (HT) with level-by-level attention pooling. Three further
baselines test whether difficulty is design-specific: a sequential
transformer (seq-T) with mean pooling, an order-free bag-of-actions MLP
(bag-MLP), and a relational graph network (R-GCN) that consumes the
typed episode graph through relation-specific message passing, included
to test whether a graph-aware model closes the compositional gap. All
reuse the same features, splits, and graphs. Numbers are means over
three to five seeds with standard deviations on the cross-subject split.
Per-level macro-F1 is reported because the action level is long-tailed.}

\rev{\textbf{Recognition across baseline families.}
Across seeds (Table~\ref{tbl:recog}), the graph-aware R-GCN is the
strongest recognizer at the HLI level ($0.754$), ahead of the
hierarchical and sequential transformers and well ahead of the
bag-of-actions model; the action level is the lowest of the four for
every baseline, consistent with its long tail. The benchmark is thus
solvable but not saturated, and discriminates between architectures
rather than being trivially easy or impossibly hard.}

\begin{table}[!htbp]
\color{black}
\footnotesize
\caption{\rev{Reference recognition on the cross-subject test split (test
macro-F1 in \%, mean $\pm$ standard deviation; HT and R-GCN over five seeds,
seq-T over three, bag-MLP over four). Four logic-free baselines from
different model families characterize difficulty; the R-GCN recognizes
best and the action level is the lowest for all
four.}}\label{tbl:recog}
\begin{tabular*}{\tblwidth}{@{}LCCCC@{}}
\toprule
Baseline & Action & Activity & LLI & HLI \\
\midrule
Hierarchical transformer & $59.0(3.4)$ & $69.5(2.2)$ & $75.7(0.8)$ & $73.2(1.1)$ \\
Sequential transformer   & $60.7(1.3)$ & $66.7(1.4)$ & $78.4(1.4)$ & $70.7(1.2)$ \\
Bag-of-actions MLP       & $44.0(4.0)$ & $50.3(3.5)$ & $63.2(5.0)$ & $56.3(2.6)$ \\
R-GCN (graph)            & $62.3(2.9)$ & $74.8(3.0)$ & $77.1(2.7)$ & $75.4(2.2)$ \\
\bottomrule
\end{tabular*}
\end{table}

\rev{\textbf{Compositional generalization.}
On the compositional held-out split, HLI macro-F1 falls by $0.13$ to
$0.17$ across all four model families and every seed
(Table~\ref{tbl:gap}); the bootstrap $95\%$ confidence interval on the
hierarchical-transformer gap over five seeds is $[0.161, 0.171]$,
excluding zero. Crucially, the graph-aware R-GCN is the strongest
recognizer yet shows the largest gap, so a model built for the typed
graph structure does not close it; the gap is a structural property of
the benchmark rather than a capacity artifact. A decomposition confirms
genuine composition rather than distribution shift: the drop is far
larger on the $1{,}854$ episodes containing a held-out
LLI$\rightarrow$HLI pairing in novel context ($0.47$ on average) than on
the $321$ episodes held out by ordering pattern alone ($0.30$), both far
above the near-zero chance level. This architecture-independent gap,
concentrated on genuine novel compositions, is the central empirical
result.}

\begin{table}[!htbp]
\color{black}
\caption{\rev{Compositional generalization: HLI macro-F1 (\%) on the standard
versus compositional held-out split (mean with standard deviation in parentheses; HT and R-GCN over five seeds, seq-T over three, bag-MLP over four). The gap is consistent across model families, and the graph-aware R-GCN does not close it.}}\label{tbl:gap}
\begin{tabular*}{\tblwidth}{@{}LCCC@{}}
\toprule
Baseline & Standard & Compositional & Gap \\
\midrule
Hierarchical transformer & $73.2(1.1)$ & $56.5(1.4)$ & $16.7(0.5)$ \\
Sequential transformer   & $70.7(1.2)$ & $56.4(1.5)$ & $14.3(0.7)$ \\
Bag-of-actions MLP       & $56.3(2.6)$ & $43.5(0.9)$ & $12.7(1.8)$ \\
R-GCN (graph)            & $75.4(2.2)$ & $58.2(3.1)$ & $17.2(1.1)$ \\
\bottomrule
\end{tabular*}
\end{table}

\rev{\textbf{Logic-violation rates as a benchmark property.}
Because the generator never enforces the Group~Y semantic rules, the
rate at which a logic-free baseline violates them measures how far the
data depart from those rules. Across seeds this held-out Group~Y rate is
low but stable for every baseline ($0.028$ to $0.042$,
Table~\ref{tbl:viol}), while the generator-shared Group~X rate is
higher, as expected. Critically, the generated ground-truth labels
themselves violate Group~Y at an intrinsic rate of $0.025$ while
satisfying Group~X at essentially zero; the Group~Y rules are therefore
genuinely independent of the generator rather than satisfied by
construction, and every model sits above this $0.025$ data floor, so the
excess is genuine prediction error rather than reproduction of the
generator. The benchmark cannot be solved by recovering the generator.}

\begin{table}[!htbp]
\color{black}
\caption{\rev{First-order-logic violation rates(\%) on the test split (mean with standard deviation in parentheses; HT and R-GCN over five seeds, seq-T over three, bag-MLP over four). Group~X rules are shared with the generator; Group~Y rules are semantic and evaluation-only. The data row is the intrinsic violation rate of the generated ground-truth labels; every model exceeds the $2.5\%$ Group~Y data floor, confirming genuine error rather than generator recovery.}}\label{tbl:viol}
\begin{tabular*}{\tblwidth}{@{}LCC@{}}
\toprule
Baseline & Group~X (shared) & Group~Y (semantic) \\
\midrule
Hierarchical transformer & $9.6(0.5)$ & $3.5(0.2)$ \\
Sequential transformer   & $10.1(0.1)$ & $3.2(0.1)$ \\
Bag-of-actions MLP       & $24.2(2.5)$ & $4.2(0.5)$ \\
R-GCN (graph)            & $9.4(0.9)$ & $2.8(0.1)$ \\
\midrule
Data (ground-truth labels) & $\approx 0\%$ & $2.5\%$ \\
\bottomrule
\end{tabular*}
\end{table}

\rev{\textbf{Order-destroying control: a generator-consistency check.}
Destroying temporal order (permuting siblings while holding every label,
performer, clip, length, and membership fixed) changes macro-F1 by less
than one standard deviation at every level and baseline
(Table~\ref{tbl:control}): a small positive drop at the lower levels
(around $0.02$ to $0.04$, where short within-action orderings carry weak
signal) and essentially zero at the intention level. The control is
therefore a generator-consistency check: the benchmark encodes order
only weakly below and not at all at the intention levels, so high-level
intention is recoverable from \emph{which} sub-behaviors co-occur rather
than their order. This order flexibility at the intention level is a
designed property of the current benchmark, not an empirical claim about
intention; the compositional held-out split, not the order control, is
the appropriate stress test for the intention levels.}

\begin{table}[!htbp]
\color{black}
\footnotesize  
\caption{\rev{Order-destroying control: drop in test macro-F1
(structured minus order-destroyed; positive means order helps), mean
$\pm$ standard deviation in parentheses over three seeds. Every value is within one
standard deviation of zero; the small positive drops at the lower levels
and the near-zero drop at the intention level show order carries only
weak signal below and none above.}}\label{tbl:control}
\begin{tabular*}{\tblwidth}{@{}LCCCC@{}}
\toprule
Baseline & Action & Activity & LLI & HLI \\
\midrule
Hierarchical transformer & $+2.9(2.0)$ & $+2.3(2.0)$ & $+4.4(4.9)$ & $+0.3(2.7)$ \\
Sequential transformer   & $+5.2(5.5)$ & $+4.0(6.1)$ & $+4.8(3.5)$ & $+1.4(2.8)$ \\
Bag-of-actions MLP       & $+1.9(2.8)$ & $+2.1(1.3)$ & $+0.9(2.6)$ & $+1.3(4.5)$ \\
\bottomrule
\end{tabular*}
\end{table}

\section{Discussion and Conclusion}\label{sec:discussion}

A benchmark-generation and evaluation framework has been presented that
synthesizes a four-level hierarchical-intention benchmark from a flat
single-label action corpus, addressing benchmark validity and
non-uniform subject coverage by design, and instantiated into a
$15{,}002$-episode benchmark. \rev{The evaluation establishes three
things. First, the compositional held-out split yields a generalization
gap of $0.13$ to $0.17$ macro-F1, consistent across four baseline
families and concentrated on genuinely novel compositions rather than
distribution shift; the graph-aware R-GCN recognizes best yet shows the
largest gap, so the gap is a structural property of the benchmark, not a
model artifact. This is the central empirical result. Second, logic-free
baselines violate the held-out semantic (Group~Y) rules above their
$0.025$ intrinsic data rate, confirming the rules are not satisfied by
construction and the benchmark resists circular shortcuts. Third, an
order-destroying control changes macro-F1 by less than seed variation at
every level, functioning as a generator-consistency check rather than
evidence about temporal order; order flexibility at the intention level
is a designed property. Overall, the novelty is not any isolated
component but their integration into a reproducible protocol that makes
hierarchy, coverage, circularity, and compositional generalization
measurable in one setting.}

\rev{A recurring concern in ontology-driven benchmarks is that the defined
hierarchy may be idiosyncratic. This concern applies equally to recorded
datasets, where scenario designers choose which activities to enact and
how to structure them. The present ontology is defined by domain experts,
released in full for inspection and critique, and supported by the
plausibility check of Section~\ref{sec:validity}. What synthesis adds over recording is
not immunity to ontology bias but transparency: the hierarchy is explicit
and modifiable rather than implicit in a recording protocol.}

Several limitations are acknowledged. The intentions are imposed by construction rather than observed, so the benchmark measures structured reasoning over a defined hierarchy rather than recovery of naturally occurring intent; the validity protocol is what makes results on it interpretable. 
The order-destroying control is conservative because short episodes contain unshuffleable single-child parents. For
two-person source actions, subject consistency is anchored to one
performer, and the action level is long-tailed, both reported
transparently. \rev{Although instantiated on one corpus, the framework
transfers to other flat single-label action corpora that satisfy three
properties: per-performer (subject) metadata, so that subject-consistent
episodes can be assembled; sufficient per-subject class coverage to form
non-degenerate eligible pools for the coverage-aware sampler; and action
granularity fine enough that classes can be composed into meaningful
higher-level units. A formal demonstration on a second source corpus,
the integration of logical constraints into models trained on the
benchmark (neuro-symbolic recognition), and enrichment of the transition
model with a directed weighted action-succession graph to strengthen
temporal signal at the action level, are left for future work.}
The intended use is controlled evaluation of hierarchical and compositional action-recognition models, not deployment for inferring real human intentions in sensitive settings.

\section*{Data Availability}
NTU RGB+D 120 licensing prohibits redistribution of source data or extracted features. The release therefore comprises the ontology, synthesis code, FOL rule set, and analysis scripts; the benchmark can be regenerated by any user with NTU RGB+D 120 access. The repository will be made available upon publication.

\section*{Declaration of generative AI and AI-assisted technologies in the manuscript preparation process}
AI tools were used to assist with drafting, editing, and analysis scripting. All content was reviewed and revised by the authors, who take full responsibility for the published work.
\printcredits

\bibliographystyle{cas-model2-names}
\bibliography{cas-refs}

\begin{thebibliography}{25}
\expandafter\ifx\csname natexlab\endcsname\relax\def\natexlab#1{#1}\fi
\providecommand{\url}[1]{\texttt{#1}}
\providecommand{\href}[2]{#2}
\providecommand{\path}[1]{#1}
\providecommand{\DOIprefix}{doi:}
\providecommand{\ArXivprefix}{arXiv:}
\providecommand{\URLprefix}{URL: }
\providecommand{\Pubmedprefix}{pmid:}
\providecommand{\doi}[1]{\href{http://dx.doi.org/#1}{\path{#1}}}
\providecommand{\Pubmed}[1]{\href{pmid:#1}{\path{#1}}}
\providecommand{\bibinfo}[2]{#2}
\ifx\xfnm\relax \def\xfnm[#1]{\unskip,\space#1}\fi
\bibitem[{Badreddine et~al.(2022)Badreddine, Garcez, Serafini and
  Spranger}]{badreddine2022logic}
\bibinfo{author}{Badreddine, S.}, \bibinfo{author}{Garcez, A.d.},
  \bibinfo{author}{Serafini, L.}, \bibinfo{author}{Spranger, M.},
  \bibinfo{year}{2022}.
\newblock \bibinfo{title}{Logic tensor networks}.
\newblock \bibinfo{journal}{Artificial Intelligence} \bibinfo{volume}{303},
  \bibinfo{pages}{103649}.
\bibitem[{Benavent-Lledo et~al.(2025)Benavent-Lledo, Mulero-P{\'e}rez,
  Ortiz-Perez, Garcia-Rodriguez and Argyros}]{benavent2025enhancing}
\bibinfo{author}{Benavent-Lledo, M.}, \bibinfo{author}{Mulero-P{\'e}rez, D.},
  \bibinfo{author}{Ortiz-Perez, D.}, \bibinfo{author}{Garcia-Rodriguez, J.},
  \bibinfo{author}{Argyros, A.}, \bibinfo{year}{2025}.
\newblock \bibinfo{title}{Enhancing action recognition by leveraging the
  hierarchical structure of actions and textual context}.
\newblock \bibinfo{journal}{Computer Vision and Image Understanding} ,
  \bibinfo{pages}{104560}.
\bibitem[{Cao et~al.(2025)Cao, Hu, Yu and Zhang}]{cao2025vision}
\bibinfo{author}{Cao, C.}, \bibinfo{author}{Hu, L.}, \bibinfo{author}{Yu, Y.},
  \bibinfo{author}{Zhang, Y.}, \bibinfo{year}{2025}.
\newblock \bibinfo{title}{Vision and intention boost large language model in
  long-term action anticipation}.
\newblock \bibinfo{journal}{arXiv preprint arXiv:2505.01713} .
\bibitem[{Cruz et~al.(2020)Cruz, Cherian, Fernando, Campbell and
  Gould}]{cruz2020inferring}
\bibinfo{author}{Cruz, R.S.}, \bibinfo{author}{Cherian, A.},
  \bibinfo{author}{Fernando, B.}, \bibinfo{author}{Campbell, D.},
  \bibinfo{author}{Gould, S.}, \bibinfo{year}{2020}.
\newblock \bibinfo{title}{Inferring temporal compositions of actions using
  probabilistic automata}, in: \bibinfo{booktitle}{Proceedings of the IEEE/CVF
  Conference on Computer Vision and Pattern Recognition Workshops}, pp.
  \bibinfo{pages}{368--369}.
\bibitem[{De~Souza et~al.(2017)De~Souza, Gaidon, Cabon and López}]{8099761}
\bibinfo{author}{De~Souza, C.R.}, \bibinfo{author}{Gaidon, A.},
  \bibinfo{author}{Cabon, Y.}, \bibinfo{author}{López, A.M.},
  \bibinfo{year}{2017}.
\newblock \bibinfo{title}{Procedural generation of videos to train deep action
  recognition networks}, in: \bibinfo{booktitle}{2017 IEEE Conference on
  Computer Vision and Pattern Recognition (CVPR)}, pp.
  \bibinfo{pages}{2594--2604}.
\newblock \DOIprefix\doi{10.1109/CVPR.2017.278}.
\bibitem[{Gebru et~al.(2021)Gebru, Morgenstern, Vecchione, Vaughan, Wallach,
  Iii and Crawford}]{gebru2021datasheets}
\bibinfo{author}{Gebru, T.}, \bibinfo{author}{Morgenstern, J.},
  \bibinfo{author}{Vecchione, B.}, \bibinfo{author}{Vaughan, J.W.},
  \bibinfo{author}{Wallach, H.}, \bibinfo{author}{Iii, H.D.},
  \bibinfo{author}{Crawford, K.}, \bibinfo{year}{2021}.
\newblock \bibinfo{title}{Datasheets for datasets}.
\newblock \bibinfo{journal}{Communications of the ACM} \bibinfo{volume}{64},
  \bibinfo{pages}{86--92}.
\bibitem[{Gill et~al.(2025)Gill, Ravichander and Marasovi{\'c}}]{gill2025has}
\bibinfo{author}{Gill, A.}, \bibinfo{author}{Ravichander, A.},
  \bibinfo{author}{Marasovi{\'c}, A.}, \bibinfo{year}{2025}.
\newblock \bibinfo{title}{What has been lost with synthetic evaluation?}
\newblock \bibinfo{journal}{arXiv preprint arXiv:2505.22830} .
\bibitem[{Hu et~al.(2020)Hu, Dong, Wang and Sun}]{hu2020heterogeneous}
\bibinfo{author}{Hu, Z.}, \bibinfo{author}{Dong, Y.}, \bibinfo{author}{Wang,
  K.}, \bibinfo{author}{Sun, Y.}, \bibinfo{year}{2020}.
\newblock \bibinfo{title}{Heterogeneous graph transformer}, in:
  \bibinfo{booktitle}{Proceedings of the web conference 2020}, pp.
  \bibinfo{pages}{2704--2710}.
\bibitem[{Kim and Linzen(2020)}]{kim2020cogs}
\bibinfo{author}{Kim, N.}, \bibinfo{author}{Linzen, T.}, \bibinfo{year}{2020}.
\newblock \bibinfo{title}{Cogs: A compositional generalization challenge based
  on semantic interpretation}, in: \bibinfo{booktitle}{Proceedings of the 2020
  conference on empirical methods in natural language processing (emnlp)}, pp.
  \bibinfo{pages}{9087--9105}.
\bibitem[{Li et~al.(2024)Li, Feng, Xu, Li, Wu, Awais, Atito and
  Kittler}]{li2024c2c}
\bibinfo{author}{Li, R.}, \bibinfo{author}{Feng, Z.}, \bibinfo{author}{Xu, T.},
  \bibinfo{author}{Li, L.}, \bibinfo{author}{Wu, X.J.}, \bibinfo{author}{Awais,
  M.}, \bibinfo{author}{Atito, S.}, \bibinfo{author}{Kittler, J.},
  \bibinfo{year}{2024}.
\newblock \bibinfo{title}{C2c: Component-to-composition learning for zero-shot
  compositional action recognition}, in: \bibinfo{booktitle}{European
  Conference on Computer Vision}, \bibinfo{organization}{Springer}. pp.
  \bibinfo{pages}{369--388}.
\bibitem[{Liu et~al.(2019)Liu, Shahroudy, Perez, Wang, Duan and
  Kot}]{liu2019ntu}
\bibinfo{author}{Liu, J.}, \bibinfo{author}{Shahroudy, A.},
  \bibinfo{author}{Perez, M.}, \bibinfo{author}{Wang, G.},
  \bibinfo{author}{Duan, L.Y.}, \bibinfo{author}{Kot, A.C.},
  \bibinfo{year}{2019}.
\newblock \bibinfo{title}{Ntu rgb+ d 120: A large-scale benchmark for 3d human
  activity understanding}.
\newblock \bibinfo{journal}{IEEE transactions on pattern analysis and machine
  intelligence} \bibinfo{volume}{42}, \bibinfo{pages}{2684--2701}.
\bibitem[{Liu et~al.(2022)Liu, Wang, Wang, Ma and Qiao}]{liu2022fineaction}
\bibinfo{author}{Liu, Y.}, \bibinfo{author}{Wang, L.}, \bibinfo{author}{Wang,
  Y.}, \bibinfo{author}{Ma, X.}, \bibinfo{author}{Qiao, Y.},
  \bibinfo{year}{2022}.
\newblock \bibinfo{title}{Fineaction: A fine-grained video dataset for temporal
  action localization}.
\newblock \bibinfo{journal}{IEEE transactions on image processing}
  \bibinfo{volume}{31}, \bibinfo{pages}{6937--6950}.
\bibitem[{Mascar{\'o} et~al.(2023)Mascar{\'o}, Ahn and
  Lee}]{mascaro2023intention}
\bibinfo{author}{Mascar{\'o}, E.V.}, \bibinfo{author}{Ahn, H.},
  \bibinfo{author}{Lee, D.}, \bibinfo{year}{2023}.
\newblock \bibinfo{title}{Intention-conditioned long-term human egocentric
  action anticipation}, in: \bibinfo{booktitle}{Proceedings of the IEEE/CVF
  Winter Conference on Applications of Computer Vision}, pp.
  \bibinfo{pages}{6048--6057}.
\bibitem[{Materzynska et~al.(2020)Materzynska, Xiao, Herzig, Xu, Wang and
  Darrell}]{materzynska2020something}
\bibinfo{author}{Materzynska, J.}, \bibinfo{author}{Xiao, T.},
  \bibinfo{author}{Herzig, R.}, \bibinfo{author}{Xu, H.},
  \bibinfo{author}{Wang, X.}, \bibinfo{author}{Darrell, T.},
  \bibinfo{year}{2020}.
\newblock \bibinfo{title}{Something-else: Compositional action recognition with
  spatial-temporal interaction networks}, in: \bibinfo{booktitle}{Proceedings
  of the IEEE/CVF conference on computer vision and pattern recognition}, pp.
  \bibinfo{pages}{1049--1059}.
\bibitem[{{\"O}zdel et~al.(2024){\"O}zdel, Rong, Albaba, Kuo, Wang and
  Kasneci}]{ozdel2024gaze}
\bibinfo{author}{{\"O}zdel, S.}, \bibinfo{author}{Rong, Y.},
  \bibinfo{author}{Albaba, B.M.}, \bibinfo{author}{Kuo, Y.L.},
  \bibinfo{author}{Wang, X.}, \bibinfo{author}{Kasneci, E.},
  \bibinfo{year}{2024}.
\newblock \bibinfo{title}{Gaze-guided graph neural network for action
  anticipation conditioned on intention}, in: \bibinfo{booktitle}{Proceedings
  of the 2024 Symposium on Eye Tracking Research and Applications}, pp.
  \bibinfo{pages}{1--9}.
\bibitem[{Piergiovanni et~al.(2020)Piergiovanni, Angelova, Toshev and
  Ryoo}]{piergiovanni2020adversarial}
\bibinfo{author}{Piergiovanni, A.}, \bibinfo{author}{Angelova, A.},
  \bibinfo{author}{Toshev, A.}, \bibinfo{author}{Ryoo, M.S.},
  \bibinfo{year}{2020}.
\newblock \bibinfo{title}{Adversarial generative grammars for human activity
  prediction}, in: \bibinfo{booktitle}{European Conference on Computer Vision},
  \bibinfo{organization}{Springer}. pp. \bibinfo{pages}{507--523}.
\bibitem[{Qi et~al.(2017)Qi, Huang, Wei and Zhu}]{qi2017predicting}
\bibinfo{author}{Qi, S.}, \bibinfo{author}{Huang, S.}, \bibinfo{author}{Wei,
  P.}, \bibinfo{author}{Zhu, S.C.}, \bibinfo{year}{2017}.
\newblock \bibinfo{title}{Predicting human activities using stochastic
  grammar}, in: \bibinfo{booktitle}{Proceedings of the IEEE International
  Conference on Computer Vision}, pp. \bibinfo{pages}{1164--1172}.
\bibitem[{Sener et~al.(2022)Sener, Chatterjee, Shelepov, He, Singhania, Wang
  and Yao}]{sener2022assembly101}
\bibinfo{author}{Sener, F.}, \bibinfo{author}{Chatterjee, D.},
  \bibinfo{author}{Shelepov, D.}, \bibinfo{author}{He, K.},
  \bibinfo{author}{Singhania, D.}, \bibinfo{author}{Wang, R.},
  \bibinfo{author}{Yao, A.}, \bibinfo{year}{2022}.
\newblock \bibinfo{title}{Assembly101: A large-scale multi-view video dataset
  for understanding procedural activities}, in: \bibinfo{booktitle}{Proceedings
  of the IEEE/CVF Conference on Computer Vision and Pattern Recognition}, pp.
  \bibinfo{pages}{21096--21106}.
\bibitem[{Todorovic(2012)}]{todorovic2012human}
\bibinfo{author}{Todorovic, S.}, \bibinfo{year}{2012}.
\newblock \bibinfo{title}{Human activities as stochastic kronecker graphs}, in:
  \bibinfo{booktitle}{European Conference on Computer Vision},
  \bibinfo{organization}{Springer}. pp. \bibinfo{pages}{130--143}.
\bibitem[{Xu et~al.(2018)Xu, Zhang, Friedman, Liang and
  Broeck}]{xu2018semantic}
\bibinfo{author}{Xu, J.}, \bibinfo{author}{Zhang, Z.},
  \bibinfo{author}{Friedman, T.}, \bibinfo{author}{Liang, Y.},
  \bibinfo{author}{Broeck, G.}, \bibinfo{year}{2018}.
\newblock \bibinfo{title}{A semantic loss function for deep learning with
  symbolic knowledge}, in: \bibinfo{booktitle}{International conference on
  machine learning}, \bibinfo{organization}{PMLR}. pp.
  \bibinfo{pages}{5502--5511}.
\bibitem[{Xu et~al.(2024)Xu, Wang, Xie, He, Zhou, Wang, Wan, Chen, Qu and
  Chu}]{xu2024logicmp}
\bibinfo{author}{Xu, W.}, \bibinfo{author}{Wang, J.}, \bibinfo{author}{Xie,
  L.}, \bibinfo{author}{He, J.}, \bibinfo{author}{Zhou, H.},
  \bibinfo{author}{Wang, T.}, \bibinfo{author}{Wan, X.}, \bibinfo{author}{Chen,
  J.}, \bibinfo{author}{Qu, C.}, \bibinfo{author}{Chu, W.},
  \bibinfo{year}{2024}.
\newblock \bibinfo{title}{Logicmp: A neuro-symbolic approach for encoding
  first-order logic constraints}, in: \bibinfo{booktitle}{International
  Conference on Learning Representations}, pp. \bibinfo{pages}{4181--4209}.
\bibitem[{Yan et~al.(2026)Yan, Zhang, Tan and Li}]{yan2026skelformer}
\bibinfo{author}{Yan, J.}, \bibinfo{author}{Zhang, X.}, \bibinfo{author}{Tan,
  C.}, \bibinfo{author}{Li, D.}, \bibinfo{year}{2026}.
\newblock \bibinfo{title}{Skelformer: An adaptive hierarchical
  transformer-based approach on skeleton graphs for human action recognition in
  video sequences}.
\newblock \bibinfo{journal}{PloS one} \bibinfo{volume}{21},
  \bibinfo{pages}{e0340390}.
\bibitem[{Ye et~al.(2025)Ye, Li, Li, Xiao and Chen}]{ye2025zero}
\bibinfo{author}{Ye, G.}, \bibinfo{author}{Li, L.}, \bibinfo{author}{Li, K.},
  \bibinfo{author}{Xiao, J.}, \bibinfo{author}{Chen, L.}, \bibinfo{year}{2025}.
\newblock \bibinfo{title}{Zero-shot compositional action recognition with
  neural logic constraints}, in: \bibinfo{booktitle}{Proceedings of the 33rd
  ACM International Conference on Multimedia}, pp. \bibinfo{pages}{3625--3634}.
\bibitem[{Zatsarynna and Gall(2023)}]{zatsarynna2023action}
\bibinfo{author}{Zatsarynna, O.}, \bibinfo{author}{Gall, J.},
  \bibinfo{year}{2023}.
\newblock \bibinfo{title}{Action anticipation with goal consistency}, in:
  \bibinfo{booktitle}{2023 IEEE International Conference on Image Processing
  (ICIP)}, \bibinfo{organization}{IEEE}. pp. \bibinfo{pages}{1630--1634}.
\bibitem[{Zhu et~al.(2025)Zhu, Zhang, Xu, Liu, Long and Wang}]{zhu2025hhgt}
\bibinfo{author}{Zhu, Q.}, \bibinfo{author}{Zhang, L.}, \bibinfo{author}{Xu,
  Q.}, \bibinfo{author}{Liu, K.}, \bibinfo{author}{Long, C.},
  \bibinfo{author}{Wang, X.}, \bibinfo{year}{2025}.
\newblock \bibinfo{title}{Hhgt: hierarchical heterogeneous graph transformer
  for heterogeneous graph representation learning}, in:
  \bibinfo{booktitle}{Proceedings of the Eighteenth ACM International
  Conference on Web Search and Data Mining}, pp. \bibinfo{pages}{318--326}.

\end{thebibliography}

\end{document}